\documentclass[conference]{IEEEtran}

\IEEEoverridecommandlockouts
\usepackage{cite}
\usepackage{amsmath,amssymb,amsfonts}
\usepackage{algorithmic}
\usepackage{graphicx}
\usepackage{hyperref}
\usepackage{textcomp}
\usepackage{xcolor}
\def\BibTeX{{\rm B\kern-.05em{\sc i\kern-.025em b}\kern-.08em
    T\kern-.1667em\lower.7ex\hbox{E}\kern-.125emX}}

\usepackage{fancyhdr}
\makeatletter
\def\ps@IEEEtitlepagestyle{%
    \fancyhf{} 
    \fancyhead[L]{2024 27th International Conference on Computer and Information Technology (ICCIT) \\ 
    20-22 December 2024, Cox’s Bazar, Bangladesh} 
    \fancyfoot[L]{979-8-3315-1909-4/24/\$31.00 ©2024 IEEE} 
}
\makeatother

\fancypagestyle{plain}{%
    \fancyhf{} 
}

\begin{document}

\title{Multi-class heart disease Detection, Classification, and Prediction using Machine Learning Models}

\author{
\IEEEauthorblockN{1\textsuperscript{st} Mahfuzul Haque}
\IEEEauthorblockA{
\textit{Dept. of Computer Science and Engineering}\\
\textit{Bangladesh Army University}\\ \textit{of Science and Technology}\\
Saidpur, Bangladesh \\
mahfuztajnim002@gmail.com}

\and
\IEEEauthorblockN{2\textsuperscript{nd} Abu Saleh Musa Miah}
\IEEEauthorblockA{
\textit{School of Computer Science} \\ \textit{and Engineering}\\
\textit{The University of Aizu}\\
Aizuwakamatsu, Japan \\
abusalehcse.ru@gmail.com}

\and
\IEEEauthorblockN{3\textsuperscript{rd} Debashish Gupta}
\IEEEauthorblockA{
\textit{Wake Forest University}\\
North Carolina, United States \\
debashisgreielts@gmail.com}
\and

\IEEEauthorblockN{4\textsuperscript{th} Md. Maruf Al Hossain Prince}
\IEEEauthorblockA{
\textit{Dept. of Computer Science} \\ \textit {and Engineering}\\
\textit{German University Bangladesh}\\
Gazipur, Bangladesh \\
marufhossain@iut-dhaka.edu}

\and
\IEEEauthorblockN{5\textsuperscript{th} Tanzina Alam}
\IEEEauthorblockA{
\textit{Dept. of Computer Science and Engineering}\\
\textit{Bangladesh Army University}\\ \textit{of Science and Technology}\\
Saidpur, Bangladesh \\
tanzinaalma95@gmail.com}
\and

\IEEEauthorblockN{6\textsuperscript{th} Nusrat Sharmin}
\IEEEauthorblockA{
\textit{Pabna Medical College}\\
Pabna, Bangladesh \\
sharminmunni01@gmail.com}
\and

\IEEEauthorblockN{7\textsuperscript{th} Mohammed Sowket Ali}
\IEEEauthorblockA{
\textit{Dept. of Computer Science and Engineering}\\
\textit{Bangladesh Army University}\\ \textit{of Science and Technology}\\
Saidpur, Bangladesh \\
sowket@gmail.com}

\and
\IEEEauthorblockN{*8\textsuperscript{th}Jungpil Shin}
\IEEEauthorblockA{
    \textit{School of Computer Science and Engineering}\\
    \textit{The University of Aizu}\\
    Aizuwakamatsu, Japan \\
    jpshin@u-aizu.ac.jp}
}

\maketitle

\begin{abstract}
Heart disease is a leading cause of premature death worldwide, particularly among middle-aged and older adults, with men experiencing a higher prevalence. According to the World Health Organization (WHO), non-communicable diseases, including heart disease, account for 25\% (17.9 million) of global deaths, with over 43,204 annual fatalities in Bangladesh. However, the development of heart disease detection (HDD) systems tailored to the Bangladeshi population remains underexplored due to the lack of benchmark datasets and reliance on manual or limited-data approaches.
This study addresses these challenges by introducing new, ethically sourced HDD dataset, BIG-Dataset and CD dataset which incorporates comprehensive data on symptoms, examination techniques, and risk factors. Using advanced machine learning techniques, including Logistic Regression and Random Forest, we achieved a remarkable testing accuracy of up to 96.6\% with Random Forest. 
The proposed AI-driven system integrates these models and datasets to provide real-time, accurate diagnostics and personalized healthcare recommendations. By leveraging structured datasets and state-of-the-art machine learning algorithms, this research offers an innovative solution for scalable and effective heart disease detection, with the potential to reduce mortality rates and improve clinical outcomes.
\end{abstract}


\section{Introduction}
Heart disease is a severe and urgent medical problem affecting a significant portion of the population \cite{who_cvd} beside the other diseases. The potential severity of delayed diagnosis makes rapid diagnosis essential \cite{aha_understanding}. Current classification approaches often fall short in practical viability, and there is a need for a comprehensive framework suitable for real-life applications in detecting and classifying this condition \cite{mayo_symptoms,kafi2022lite_kidney_miah,miah2021alzheimer,hassan2024residual_miah_alzh}.

To address this problem, a revolutionary software solution is being developed to help healthcare practitioners identify early-stage cardiac diseases by leveraging computer technology and machine learning approaches \cite{harvard_truth}. Early identification of heart disease by evaluating a person's present cardiac difficulties and risk factors promises to reduce mortality rates \cite{johns_hopkins_diagnosis}. Given the complexity of heart disease detection and classification, existing approaches that rely on pre-processed datasets with few variables must be updated \cite{bhf_types}. A comprehensive examination, including general assessments, systemic evaluations of symptoms, risk factors and detailed investigative diagnoses are required due to the disease's complexity \cite{esc_guidelines}.

As discussed in \cite{laks}, feature selection techniques are critical in improving heart disease prediction. These techniques help identify the most relevant features from a large set of available data, thus improving the model's performance and reducing overfitting.

Standard datasets often lack applicability and relevance for real-world implementation, making precise detection and classification challenging \cite{pmc_data_augmentation}. Heart disease identification heavily relies on symptoms, risk factors, and specific investigative data, which are frequently missing from traditional datasets dominated by synthetic data \cite{google_scholar_ml}. This work, in contrast, presents a meticulously constructed structured dataset for disease categorization, consisting of over 45,000 accurate patient records from government hospitals, diagnostic centers, and online archives, supplemented with ongoing data augmentation \cite{ieee_algorithms}. This extensive dataset is poised to improve model training, resulting in consistently more accurate predictions \cite{jmir_ensemble}.

The paper's groundbreaking paradigm enables individualized disease classification and prediction by integrating cardiac disease diagnosis, categorization, and prognosis \cite{bioinformatics_integrative}. The creation of specialized datasets like HDD (Heart Disease Detection), and CD (Combined Dataset), as well as a sophisticated model using classifiers from Logistic Regression to ensemble model like Random Forest, highlights the potential for precise predictions \cite{nature_high_accuracy}.

In summary, in this paper, we make the following contributions:

\begin{itemize}
    \item We newly collected HDD and CD datasets from two different sources and classified them into different classes.
    \item To compare the effectiveness of our suggested strategies, we carried out an experimental study.
    \item We performed experimental evaluations to compare our proposed models with existing models.
\end{itemize}

This paper, divided into six chapters in which literature review is covered in \ref{LR}, dataset description in \ref{DS}, proposed methodology in \ref{PM}, experimental results in \ref{ER}, and conclusion in \ref{CON}.

\section{Literature Review} \label{LR}
Machine learning models are being used in various domains for their excellence \cite{hasan2024firelite, miah2022bensignnet, miah2024hand_multiculture}. There are many researches also have been done to develop various human disease recognition system using various machine learning and deep learning approaches \cite{hassan2024residual,miah2022movie_miah}. 
Lakshmanaro et al. used feature selection and ensemble learning techniques to predict heart disease \cite{Lakshmanarao2021}. They have used ensemble learning effectively. They have also successfully used an ensemble model to forecast cardiac disease. Beulah C. et al \cite{Christian2018} developed the ensemble classification to increase the accuracy of less robust algorithms that integrate multiple different classifiers. Jaymin Patel compares various classification methods for decision trees in his research \cite{Chang2015}.  Rani et al. proposed a hybrid system using a Support Vector Machine, Naive Bayes, Logical Regression, Random Forest, and AdaBoost classifiers in the final stage of the system's development, as Pooja Rani showed \cite{Rani2016}. This one's accuracy of 86.6\% outperforms other algorithms for predicting heart disease discovered in the scientific literature.
Lerina Aversano et al. employed machine learning for the early prediction of heart disease, showing promising outcomes in early diagnosis \cite{Aversano2022}. Dimitris Bertsimas et al. demonstrated the effectiveness of machine learning for real-time heart disease prediction \cite{Bertsimas2021}.

Additionally, Richard Raymond Bomford and Adair Stuart Masons clinical methods have laid the foundation for many modern diagnostic techniques \cite{Bomford1975}. A. H. Chen et al. developed HIPS Heart Disease Prediction System, an early attempt at integrating data mining and heart disease prediction \cite{Chen2011}. Abderrahmane Ed-Daoudy and Khalil Maalmi applied real-time machine learning techniques to big data for early detection of heart disease \cite{EdDaoudy2019}. Aditi Gavhane et al. predicted heart disease using various machine learning algorithms, showcasing the diverse applicability of these techniques \cite{Gavhane2018}.

Furthermore, M. Akhil Jabbar et al. explored lazy associative classification for heart disease prediction, providing insights into alternative classification methods \cite{Jabbar2013}. Pahulpreet Singh Kohli and Shriya Arora applied machine learning in disease prediction, contributing to the growing field of AI in healthcare \cite{Kohli2018}. Jian Ping Li et al. developed a heart disease identification method using machine learning classification in e-healthcare, reflecting the integration of AI in healthcare systems \cite{Li2020}. Senthilkumar Mohan et al. effectively used hybrid machine learning techniques for heart disease prediction, highlighting the benefits of combining multiple algorithms \cite{Mohan2019}.
Lastly, the study by Anjan Nikhil Repaka, Sai Deepak Ravikanti, and Ramya G Franklin focused on using naive Bayesian methods for heart disease prediction, adding to the array of predictive models available \cite{Repaka2019}. Priyanka S. Single et al. reviewed methodologies and techniques for heart disease classification and prediction, providing a comprehensive overview of existing approaches \cite{Single2020}. Vijeta Sharma et al. examined machine learning techniques for heart disease prediction, reinforcing the importance of continuous innovation in this field \cite{Sharma2020}. Archana Singh and Rakesh Kumar compared various machine learning algorithms for heart disease prediction, offering valuable insights into their comparative performance \cite{Singh2020}. Jagdeep Singh et al. explored associative classification for heart disease prediction, presenting another dimension of classification techniques \cite{Singh2016}. J. Thomas and R Theresa Prince utilized data mining techniques for heart disease prediction, contributing to the expanding literature on data-driven healthcare solutions \cite{Thomas2016}.

\section{Dataset} \label{DS}
The dataset was collected from government hospitals, diagnostic centers, and validated online repositories. Data from 45,779 participants (aged 3-17) were used, with 5,218 children diagnosed with ADHD. The data are binary (1 for Yes, 0 for No) and identifies symptoms like chest pain, shortness of breath, and risk factors. The dataset encompasses both affected and non-affected individuals. Ethical clearance for data collection was obtained from the Ethical Review Board of the respective institutions.

The dataset can be accessed from Kaggle:  
\textbf{Dataset Link}: \href{https://www.kaggle.com/datasets/mahfuzulhaquetajnim/heart-disease-2024/data}{Heart Disease Dataset}.
\begin{table*}[tb]
\centering
\caption{Health Assessment Questions}
\label{table:health_assessment}
\begin{tabular}{|l|l|l|l|}
\hline
\textbf{General Factor Questions} & \textbf{Categories} & \textbf{General Factor Questions} & \textbf{Categories} \\ \hline
Do you have Chest pain & Yes and No & Do you have Missing heartbeat rhythm/abnormal rhythm & Yes and No \\ \hline
Do you have Shortness of breath & Yes and No & Do you have Need pillow or prefer to sleep in chair & Yes and No \\ \hline
Do you have Heaviness or tightness & Yes and No & Do you have Syncopal attack & Yes and No \\ \hline
Do you have Radiation into arms neck and jaw & Yes and No & Do you have Debilitation & Yes and No \\ \hline
Do you have Congestion or burning & Yes and No & Do you have fever & Yes and No \\ \hline
Do you have Abnormal breathing & Yes and No & Do you have Clubbing & Yes and No \\ \hline
Do you have Losing flat cause & Yes and No & Do you have Rash & Yes and No \\ \hline
Do you have Decrease of strength & Yes and No & \textbf{Risk Factor Questions} & \\ \hline
Do you have Trouble with balance & Yes and No & Do you have Smoking & Yes and No \\ \hline
Do you have Heart rate normal or faster & Yes and No & Do you have Hypertension & Yes and No \\ \hline
Do you have Trouble with swallowing & Yes and No & Do you have Hypercholesterolemia & Yes and No \\ \hline
Do you have Low blood pressure & Yes and No & Do you have Myocardial infarction & Yes and No \\ \hline
\end{tabular}
\end{table*}

\subsection{HDD Dataset}
HDD stands for Heart Disease Detection, we have used heart disease symptoms, examine techniques, risk factors, investigation, and diagnosis information mentioned in \cite{penman2022davidson} and \cite{glynn2022hutchison}. Nineteen symptoms frequently occur across twenty-seven heart diseases. Risk factors are also present. This dataset exclusively represents patterns associated with heart disease patients. It can identify all twenty-seven heart disease categories. Additionally, the dataset allows for binary classification of heart disease for atypical samples. 
\begin{figure}[tbp]
\centerline{\includegraphics[width=0.72\linewidth]{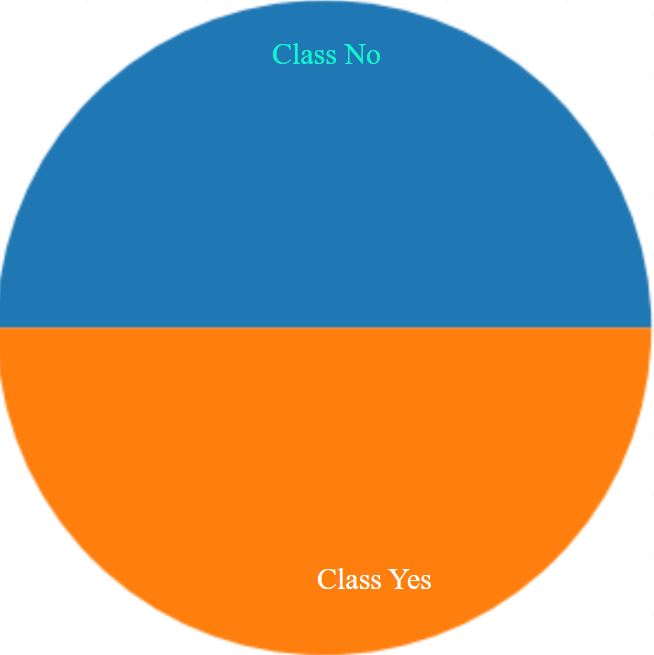}}
\caption{Distribution of the ratio of the class labels for the BidD dataset.}
\label{class_dist_bigD}
\end{figure}

\subsection{BIG-Dataset}
In 2022, over 1700 people datasets with 15 attributes were collected from hospitals and 1200 from online sources. Non-affected users' data is stored in the BIG-D dataset. The sample of this dataset is shown in Table \ref{data:bigd}

\subsection{CD Dataset}
\begin{figure}[tbp]
\centerline{\includegraphics[width=1.08\linewidth]{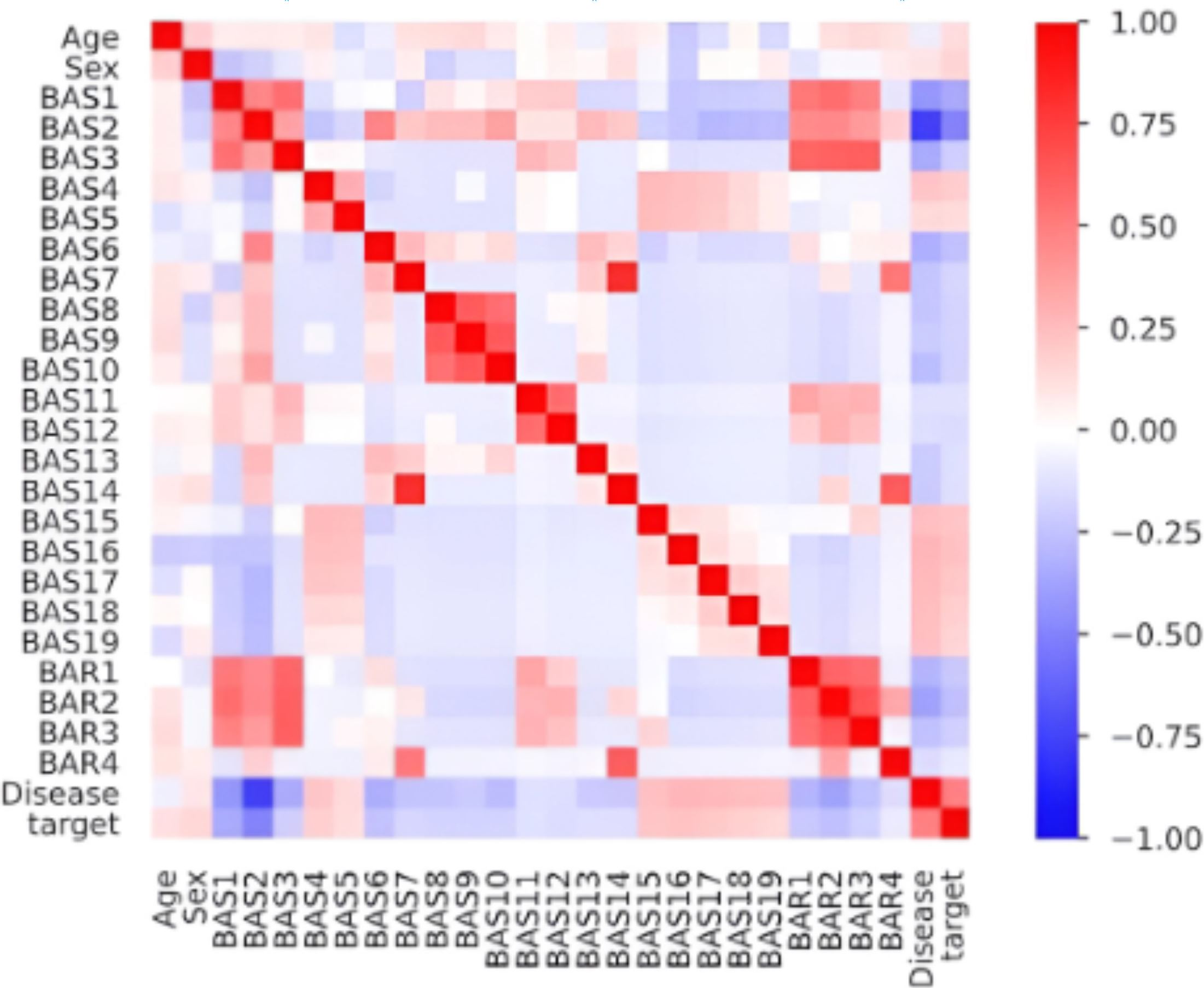}}
\caption{Heatmap of the correlation matrix for the CD dataset features.}
\label{correlation}
\end{figure}

CD stands for the combined dataset. We named this dataset according to its action. This dataset is developed from combining the HDD and BIG dataset. In this dataset, data for affected people and non-affected people are combined. We have developed this dataset to train advanced models for better accuracy and to make them usable in advanced technology.  This dataset consisted of two labels, including a normal person and a person with heart disease. According to the dataset HDD, which was created with 27 diseases, there are 19 symptoms and four risk factors. The 19 symptoms have two levels: some are major symptoms, and some are minor symptoms. The correlation matrix of the features are shown in Fig. \ref{correlation}. 

\begin{table*}[ht]
\centering
\caption{Example of the BigD Dataset} \label{data:bigd}
\begin{tabular}{|c|c|c|c|c|c|c|c|c|c|c|c|c|c|c|c|}
\hline
no&sex & age & Smkr & oldpeak & fbs & cp & restecg & exang & chol & trestbps & diaBP & BMI & heartRate & glucose & target \\
\hline
1 & 1 & 39 & 0 & 0 & 0 & 3 & 0 & 0 & 195 & 106.0 & 70.0 & 26.97 & 80 & 77 & 0 \\
2 & 0 & 46 & 0 & 0 & 0 & 2 & 0 & 0 & 250 & 121.0 & 81.0 & 28.73 & 95 & 76 & 0 \\
3 & 1 & 48 & 1 & 20 & 0 & 1 & 0 & 0 & 245 & 127.5 & 80.0 & 25.34 & 75 & 70 & 0 \\
4 & 0 & 61 & 1 & 30 & 0 & 1 & 1 & 0 & 225 & 150.0 & 95.0 & 28.58 & 65 & 103 & 1 \\
5 & 0 & 46 & 1 & 23 & 0 & 0 & 0 & 0 & 285 & 130.0 & 84.0 & 23.10 & 85 & 85 & 0 \\
\hline
\end{tabular}

\end{table*}

\subsection{Dataset Limitations}
While comprehensive, the dataset exhibits certain limitations, such as a bias toward specific population groups, which could potentially impact generalization to other demographics. Future studies should consider incorporating more diverse data.

\begin{figure}[tbp]
\centerline{\includegraphics[width=1.0\linewidth]{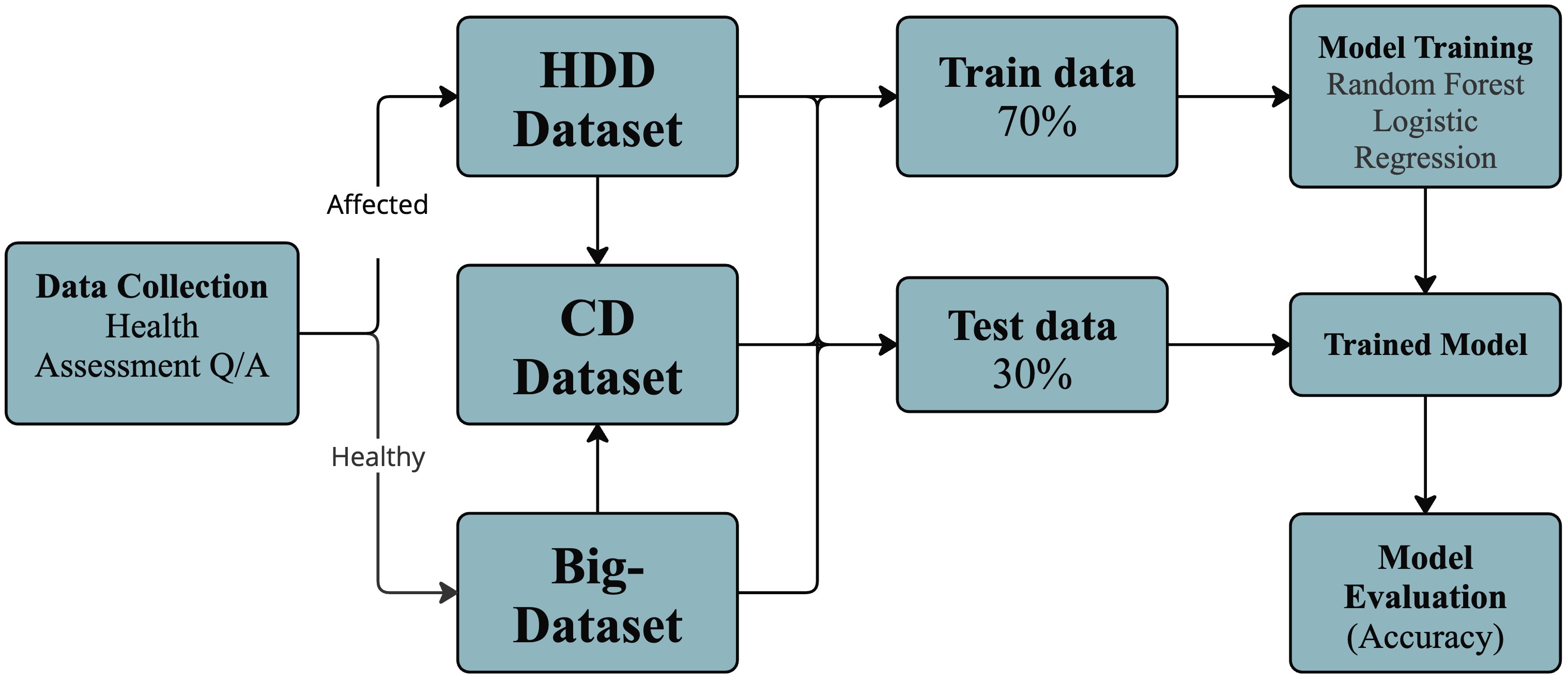}}
\caption{Workflow of the proposed model and evaluation pipeline.}
\label{flowchart}
\end{figure}

\section{Proposed Methodology} \label{PM}
In this study, we introduced a new heart disease dataset and a model for classification of heart disease. Unlike previous research using existing datasets, our framework is unique, utilizing real-life data from government hospitals, diagnostic clinics, and online sources. The workflow of our proposed methodology is shown in Fig. \ref{flowchart}


\subsection{Feature Engineering}

Features were selected based on clinical significance, including 19 symptoms and 4 risk factors based on \cite{penman2022davidson,glynn2022hutchison}. Logistic Regression was chosen for its interpretability, while Random Forest was employed for its ability to handle complex, non-linear relationships. The ensemble approach of random forest mitigates overfitting and ensures robust predictions.

\subsection{Classification}
\subsubsection{Logistic Regression}
In the study, we employed logistic regression to classify the disease and normal people, as well as the severity classification of the patient. 
Logistic regression is a supervised classification algorithm commonly used for predictive analysis based on probability. This technique evaluates the relationship between the dependent variable and independent variables or risk factors by calculating probabilities using the logistic function, also known as the sigmoid function \cite{joy2020multiclass}.
The logistic function is expressed as:
\begin{equation}
h(x) = \frac{1}{1 + e^{-(\beta_0 + \beta_1 x_1 + \beta_2 x_2 + \cdots + \beta_n x_n)}}
\end{equation}

Where \( h(x) \) represents the probability that the dependent variable is 1 (e.g., the presence of heart disease), \( \beta_0 \) is the intercept, and \( \beta_1, \beta_2, \ldots, \beta_n \) are the coefficients of the independent variables \( x_1, x_2, \ldots, x_n \).

The logistic function constrains the output between 0 and 1, which is essential for binary classification problems.
Logistic regression is particularly suitable for predicting binary outcomes (e.g., diseased vs. not diseased). The primary applications of logistic regression include prediction and estimating the probability of success. In this context, logistic regression reveals a statistically significant link between the risk factors and the likelihood of developing heart disease.

The cost function used in logistic regression is derived from the sigmoid function and is expressed as:

\begin{equation}
J(\beta) = - \frac{1}{m} \sum_{i=1}^{m} \left[ y_i \log(h(x_i)) + (1 - y_i) \log(1 - h(x_i)) \right]
\end{equation}

where \( J(\beta) \) is the cost function, \( m \) is the number of training examples, \( y_i \) is the actual output for the \( i \)-th example, and \( h(x_i) \) is the predicted probability for the \( i \)-th example.

\subsubsection{Random Forest}
In the study, we also employed Random Forest to classify the disease and normal people, as well as the severity classification of the patient.
This algorithm designed to mitigate the overfitting issue commonly associated with single decision trees. Single decision trees often learn from only one pathway of decisions, resulting in poor generalisation to new datasets \cite{miah2022eeg,miah2019motor,miah2022movie_miah}. Random Forest addresses this by constructing multiple clusters of decision trees with controlled variation, enhancing overall prediction accuracy. The Random Forest algorithm operates by merging random selections of input features and using Breimans bagging (Bootstrap Aggregating) sampling techniques. The bagging algorithm improves robustness by drawing observations with replacements from the training data and randomly splitting nodes to achieve the best split within a random subset of features. The decision function in Random Forest can be expressed as:
\begin{equation}
\hat{y} = \text{mode}\{h_1(x), h_2(x), \ldots, h_B(x)\}
\end{equation}
where \( \hat{y} \) is the predicted class, \( h_i(x) \) is the prediction from the \(i\)-th decision tree, and \( B \) is the total number of trees in the forest.
In our case, we optimized the hyperparameters of the Random Forest algorithm using a randomized search algorithm with cross-validation. This process involved tuning parameters such as the number of trees, maximum depth, and the number of features considered for splitting at each node to achieve the best performance on the validation set. The Random Forest can also help reduce the noise and overfitting of the system \cite{Breiman2001}.

\section{Experimental Results} \label{ER}
We used Random Forest and Logistic Regression on the HDD, BIG dataset and CD datasets. 

\subsection{Performance With HDD dataset and BigD Dataset}
Table \ref{tab_hdd_dataset} demonstrates the performance accuracy of the HDD dataset using different models. Random Forest achieved 92.77\% training accuracy and 91.90\% testing accuracy, while Logistic Regression achieved 95.00\% training accuracy and 93.87\% testing accuracy. Besides, the Random forest for the BigD dataset shows  90.80\% accuracy. 

\begin{table}[tb]
\caption{Accuracy with different models on HDD dataset}
\centering
\begin{tabular}{|p{2cm}|p{1cm}|p{1cm}|p{1cm}|p{1cm}|}
\hline
Algorithm & HDD Train Acc. & HDD Test Acc. & BigD Train Acc. & BigD Test Acc.  \\
\hline
Logistic Regression & 95.00\% & 93.87& - & - \\
\hline
Random Forest & 92.77\% & 91.90\%& - & 90.80\% \\
\hline

\end{tabular}
\label{tab_hdd_dataset}
\end{table}

\subsection{Performance With CD Dataset}
Similarly, Table \ref{tab_CD_dataset} demonstrates the performance accuracy of the CD dataset, with Random Forest achieving 99.21\% training accuracy and 96.66\% testing accuracy, and Logistic Regression achieving 98.66\% training accuracy and 95.67\% testing accuracy.

\begin{table}[tb]
\caption{Accuracy with different models on CD dataset}
\begin{center}
\begin{tabular}{|l|l|l|}
\hline
Algorithm &Train Accuracy & Test Accuracy \\
\hline
Logistic Regression & 98.66 & 95.67 \\
\hline
Random Forest (Proposed) & 99.21 &96.66 \\
\hline
\end{tabular}
\end{center}
\label{tab_CD_dataset}
\end{table}

\subsection{Comparative Analysis}

The proposed model achieves a testing accuracy of 96.66\%, outperforming existing studies. Table~\ref{tab:comparison} compares our results with prior research:

\begin{table}[tb]
\centering
\caption{Comparison of Accuracy with Prior Studies for CD}
\label{tab:comparison}
\begin{tabular}{|l|l|c|}
\hline
Study &Methodology & Accuracy (\%) \\ \hline
Lakshmanaro et al. & Ensemble Learning & 88.78 \\ \hline
Dixit et al. \cite{dixi} & Particle Swarm Optimization & 86.60 \\ \hline
Proposed Model & Random Forest & 96.66 \\ \hline
\end{tabular}
\end{table}

\subsection{Discussion}
Our main contribution to the study is creating real-world heart disease datasets, namely HDD, BIG, and CD.

The HDD dataset, designed for recognizing patterns in 27 heart disease categories, achieved high accuracy, with Logistic Regression at 93.87\% and Random Forest at 91.90\%.  The BIG dataset, including unaffected individuals, supports binary classification. Random Forest achieved 90.80\% accuracy, slightly lower than HDD, likely due to added complexity from non-affected samples. This highlights BIG’s value in real-world scenarios with healthy populations.

The CD dataset, combining affected and unaffected cases, achieved the highest performance, with Random Forest at 96.66\% and Logistic Regression at 95.67\%. Its balanced representation enables Random Forest to capture intricate interactions, making it ideal for both binary and multi-class tasks.

\begin{figure}[tbp]
\centerline{\includegraphics[width=0.7\linewidth]{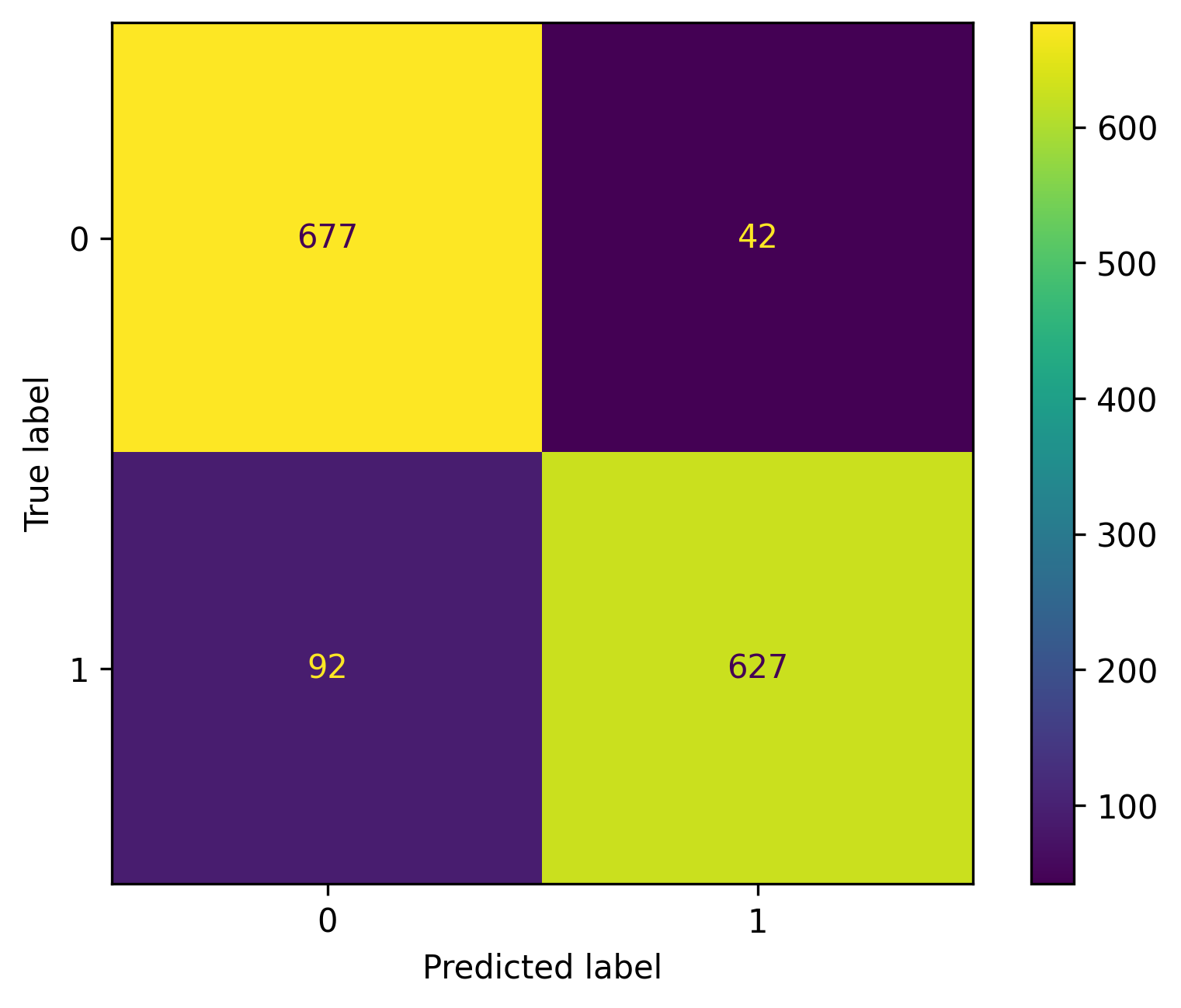}}
\caption{Performance Analysis of Classification Model.}
\label{dec}
\end{figure}

Across all datasets, Logistic Regression performed well with structured data, while Random Forest excelled in handling complexity, emphasizing the importance of matching models to dataset characteristics.The performance analysis of classification model are shown in Fig.\ref{dec}. 

\section{Conclusion} \label{CON}
This research assessed machine learning algorithms for cardiac disease identification utilizing the HDD, BIG, and CD datasets. 

These findings underscore the importance of dataset diversity in enhancing model efficacy. The HDD dataset is superior for disease-specific classification, while the BIG and CD datasets facilitate wider real-world applicability. Random Forest exhibited versatility in handling complex datasets, rendering it appropriate for clinical applications.

Future research may augment these datasets with supplementary features and investigate explainability frameworks to improve diagnostic precision and reliability. This study advances the creation of scalable, dependable instruments for cardiac disease detection.

\bibliographystyle{IEEEtran}
\bibliography{References}

\begin{thebibliography}{10}
\providecommand{\url}[1]{#1}
\csname url@samestyle\endcsname
\providecommand{\newblock}{\relax}
\providecommand{\bibinfo}[2]{#2}
\providecommand{\BIBentrySTDinterwordspacing}{\spaceskip=0pt\relax}
\providecommand{\BIBentryALTinterwordstretchfactor}{4}
\providecommand{\BIBentryALTinterwordspacing}{\spaceskip=\fontdimen2\font plus
\BIBentryALTinterwordstretchfactor\fontdimen3\font minus \fontdimen4\font\relax}
\providecommand{\BIBforeignlanguage}[2]{{%
\expandafter\ifx\csname l@#1\endcsname\relax
\typeout{** WARNING: IEEEtran.bst: No hyphenation pattern has been}%
\typeout{** loaded for the language `#1'. Using the pattern for}%
\typeout{** the default language instead.}%
\else
\language=\csname l@#1\endcsname
\fi
#2}}
\providecommand{\BIBdecl}{\relax}
\BIBdecl

\bibitem{who_cvd}
{World Health Organization}, ``Cardiovascular diseases (cvds),'' 2021, retrieved from \url{https://www.who.int/news-room/fact-sheets/detail/cardiovascular-diseases-(cvds)}.

\bibitem{aha_understanding}
{American Heart Association}, ``Understanding heart disease,'' 2021, retrieved from \url{https://www.heart.org/en/health-topics/consumer-healthcare}.

\bibitem{mayo_symptoms}
{Mayo Clinic}, ``Heart disease - symptoms and causes,'' 2021, retrieved from \url{https://www.mayoclinic.org/diseases-conditions/heart-disease/symptoms-causes/syc-20353118}.

\bibitem{kafi2022lite_kidney_miah}
H.~M. Kafi, A.~S.~M. Miah, J.~Shin, and M.~N. Siddique, ``A lite-weight clinical features based chronic kidney disease diagnosis system using 1d convolutional neural network,'' in \emph{2022 International Conference on Advancement in Electrical and Electronic Engineering (ICAEEE)}.\hskip 1em plus 0.5em minus 0.4em\relax IEEE, 2022, pp. 1--5.

\bibitem{miah2021alzheimer}
A.~S.~M. Miah, M.~Mamunur~Rashid, M.~Redwanur~Rahman, M.~Tofayel~Hossain, M.~Shahidujjaman~Sujon, N.~Nawal, M.~Hasan, and J.~Shin, ``Alzheimer’s disease detection using cnn based on effective dimensionality reduction approach,'' in \emph{Intelligent Computing and Optimization: Proceedings of the 3rd International Conference on Intelligent Computing and Optimization 2020 (ICO 2020)}.\hskip 1em plus 0.5em minus 0.4em\relax Springer, 2021, pp. 801--811.

\bibitem{hassan2024residual_miah_alzh}
N.~Hassan, A.~S. Musa~Miah, and J.~Shin, ``Residual-based multi-stage deep learning framework for computer-aided alzheimer’s disease detection,'' \emph{Journal of Imaging}, vol.~10, no.~6, p. 141, 2024.

\bibitem{harvard_truth}
{Harvard Health Publishing}, ``The truth about heart disease,'' 2021, retrieved from \url{https://www.health.harvard.edu/topics/heart-health}.

\bibitem{johns_hopkins_diagnosis}
{Johns Hopkins Medicine}, ``Heart disease: Diagnosis and tests,'' 2021, retrieved from \url{https://www.hopkinsmedicine.org/health/conditions-and-diseases/heart-disease}.

\bibitem{bhf_types}
{British Heart Foundation}, ``Understanding the different types of heart disease,'' 2021, retrieved from \url{https://www.bhf.org.uk/informationsupport/conditions/heart-diseases}.

\bibitem{esc_guidelines}
{European Society of Cardiology}, ``Esc guidelines for the diagnosis and treatment of acute and chronic heart failure,'' 2021, retrieved from \url{https://www.escardio.org/Guidelines/Clinical-Practice-Guidelines/Acute-and-Chronic-Heart-Failure}.

\bibitem{laks}
A.~Lakshmanarao, A.~Srisaila, and T.~S.~R. Kiran, ``Heart disease prediction using feature selection and ensemble learning techniques,'' \emph{Proc. Int. Conf. Intell. Commun. Technol.}, pp. 994--998, 2021.

\bibitem{pmc_data_augmentation}
{PubMed Central}, ``Data augmentation in heart disease detection,'' 2021, retrieved from \url{https://www.ncbi.nlm.nih.gov/pmc/articles/PMC7583211/}.

\bibitem{google_scholar_ml}
{Google Scholar}, ``Machine learning approaches for heart disease detection,'' 2021, retrieved from \url{https://scholar.google.com/scholar?q=machine+learning+heart+disease+detection}.

\bibitem{ieee_algorithms}
{IEEE Xplore}, ``Advanced algorithms for heart disease prediction,'' 2021, retrieved from \url{https://ieeexplore.ieee.org/document/8888660}.

\bibitem{jmir_ensemble}
{Journal of Medical Internet Research}, ``Heart disease prediction using ensemble learning,'' 2021, retrieved from \url{https://www.jmir.org/2021/6/e24845/}.

\bibitem{bioinformatics_integrative}
{Bioinformatics}, ``Integrative approaches for heart disease classification,'' 2021, retrieved from \url{https://academic.oup.com/bioinformatics/article/37/7/898/6024632}.

\bibitem{nature_high_accuracy}
{Nature Communications}, ``High accuracy heart disease detection models,'' \emph{Scientific reports}, 2021, retrieved from \url{https://www.nature.com/articles/s41467-021-23887-3}.

\bibitem{hasan2024firelite}
M.~Hasan, M.~M. A.~H. Prince, M.~S. Ansari, S.~Jahan, A.~S.~M. Miah, and J.~Shin, ``Firelite: Leveraging transfer learning for efficient fire detection in resource-constrained environments,'' \emph{arXiv preprint arXiv:2409.20384}, 2024.

\bibitem{miah2022bensignnet}
A.~S.~M. Miah, J.~Shin, M.~A.~M. Hasan, and M.~A. Rahim, ``Bensignnet: Bengali sign language alphabet recognition using concatenated segmentation and convolutional neural network,'' \emph{Applied Sciences}, vol.~12, no.~8, p. 3933, 2022.

\bibitem{miah2024hand_multiculture}
A.~S.~M. Miah, M.~A.~M. Hasan, Y.~Tomioka, and J.~Shin, ``Hand gesture recognition for multi-culture sign language using graph and general deep learning network,'' \emph{IEEE Open Journal of the Computer Society}, 2024.

\bibitem{hassan2024residual}
N.~Hassan, A.~S. Musa~Miah, and J.~Shin, ``Residual-based multi-stage deep learning framework for computer-aided alzheimer’s disease detection,'' \emph{Journal of Imaging}, vol.~10, no.~6, p. 141, 2024.

\bibitem{miah2022movie_miah}
A.~S.~M. Miah, J.~Shin, M.~A.~M. Hasan, M.~K.~I. Molla, Y.~Okuyama, and Y.~Tomioka, ``Movie oriented positive negative emotion classification from eeg signal using wavelet transformation and machine learning approaches,'' in \emph{2022 IEEE 15th international symposium on embedded multicore/many-core systems-on-chip (MCSoC)}.\hskip 1em plus 0.5em minus 0.4em\relax IEEE, 2022, pp. 26--31.

\bibitem{Lakshmanarao2021}
A.~Lakshmanarao, A.~Srisaila, and T.~S.~R. Kiran, ``Heart disease prediction using feature selection and ensemble learning techniques,'' in \emph{2021 Third International Conference on Intelligent Communication Technologies and Virtual Mobile Networks (ICICV)}, 2021, pp. 994--998, 10.

\bibitem{Christian2018}
C.~B.~C. Latha, ``Ensemble classification by investigating a method to improve the accuracy of weak algorithms that combine multiple classifiers,'' in \emph{2018 International Conference on Smart Technologies for Smart Nation}, 2018, pp. 51--68, 23.

\bibitem{Chang2015}
V.~Chang, ``Comparing different decision tree classification algorithms using weka for better performance in heart disease diagnosis,'' in \emph{ICWE 2015}, 2015, pp. 51--68, 24.

\bibitem{Rani2016}
P.~Rani, ``Development of a proposed hybrid system using support vector machine, naive bayes, logistic regression, random forest, and adaboost classifier,'' in \emph{2016 International Conference on Smart Technologies for Smart Nation}, 2016, pp. 129--338, 25.

\bibitem{Aversano2022}
L.~Aversano, M.~L. Bernardi, M.~Cimitile, M.~Iammarino, D.~Montano, and C.~Verdone, ``Using machine learning for early prediction of heart disease,'' in \emph{2022 IEEE International Conference on Evolving and Adaptive Intelligent Systems (EAIS)}, 2022, pp. 1--8, 2.

\bibitem{Bertsimas2021}
D.~Bertsimas, L.~Mingardi, and B.~Stellato, ``Machine learning for real-time heart disease prediction,'' \emph{IEEE Journal of Biomedical and Health Informatics}, vol.~25, no.~9, pp. 3627--3637, 2021, 3.

\bibitem{Bomford1975}
R.~R. Bomford and A.~S. Mason, \emph{Hutchison's clinical methods}.\hskip 1em plus 0.5em minus 0.4em\relax Bailliere Tindall, 1975, 4.

\bibitem{Chen2011}
A.~H. Chen, S.~Y. Huang, P.~S. Hong, C.~H. Cheng, and E.~J. Lin, ``Hips: Heart disease prediction system,'' in \emph{2011 Computing in Cardiology}, 2011, pp. 557--560, 5.

\bibitem{EdDaoudy2019}
A.~Ed-Daoudy and K.~Maalmi, ``Real-time machine learning for early detection of heart disease using a big data approach,'' in \emph{2019 International Conference on Wireless Technologies, Embedded and Intelligent Systems (IEEE)}, 2019, pp. 1--5, 6.

\bibitem{Gavhane2018}
A.~Gavhane, G.~Kokkula, I.~Pandya, and K.~Devadkar, ``Prediction of heart disease using machine learning,'' in \emph{2018 Second International Conference on Electronics, Communication and Aerospace Technology (ICEEE)}, 2018, pp. 1275--1278, 7.

\bibitem{Jabbar2013}
M.~A. Jabbar, B.~L. Deekshatulu, and P.~Chandra, ``Heart disease prediction using lazy associative classification,'' in \emph{2013 International Multi-Conference on Automation, Computing, Communication, Control and Compressed Sensing (iMac4s)}, 2013, pp. 40--46, 8.

\bibitem{Kohli2018}
P.~S. Kohli and S.~Arora, ``Application of machine learning in disease prediction,'' in \emph{2018 4th International Conference on Computing Communication and Automation (IEEE)}, 2018, pp. 1--4, 9.

\bibitem{Li2020}
J.~P. Li, A.~U. Haq, S.~U. Din, J.~Khan, A.~Khan, and A.~Saboor, ``Heart disease identification method using machine learning classification in e-healthcare,'' \emph{IEEE Access}, vol.~8, pp. 107\,562--107\,582, 2020, 11.

\bibitem{Mohan2019}
S.~Mohan, C.~Thirumalai, and G.~Srivastava, ``Effective heart disease prediction using hybrid machine learning techniques,'' \emph{IEEE Access}, vol.~7, pp. 81\,542--81\,554, 2019, 12.

\bibitem{Repaka2019}
A.~N. Repaka, S.~D. Ravikanti, and R.~G. Franklin, ``Design and implement heart disease prediction using natives bayesian,'' in \emph{2019 3rd International Conference on Trends in Electronics and Informatics (IEEE)}, 2019, pp. 292--297, 14.

\bibitem{Single2020}
P.~S. Single, R.~M. Goudar, and A.~Bhutto, ``Methodologies and techniques for heart disease classification and prediction,'' in \emph{2020 11th International Conference on Computing, Communication and Networking Technologies (IEEE)}, 2020, pp. 1--6, 15.

\bibitem{Sharma2020}
V.~Sharma, S.~Yadav, and M.~Gupta, ``Heart disease prediction using machine learning techniques,'' in \emph{2020 2nd International Conference on Advances in Computing, Communication Control and Networking (IEEE)}, 2020, pp. 177--181, 16.

\bibitem{Singh2020}
A.~Singh and R.~Kumar, ``Heart disease prediction using machine learning algorithms,'' in \emph{2020 International Conference on Electrical and Electronics Engineering (ICE3)}, 2020, 17.

\bibitem{Singh2016}
J.~Singh, A.~Kamra, and H.~Singh, ``Prediction of heart diseases using associative classification,'' in \emph{2016 5th International Conference on Wireless Networks and Embedded Systems (ICEEE)}, 2016, pp. 1--7, 18.

\bibitem{Thomas2016}
J.~Thomas and R.~T. Prince, ``Human heart disease prediction system using data mining techniques,'' in \emph{2016 International Conference on Circuit, Power and Computing Technologies (ICEEE)}, 2016, pp. 1--5, 19.

\bibitem{penman2022davidson}
I.~D. Penman, S.~H. Ralston, M.~W. Strachan, and R.~Hobson, \emph{Davidson's Principles and Practice of Medicine E-Book: Davidson's Principles and Practice of Medicine E-Book}.\hskip 1em plus 0.5em minus 0.4em\relax Elsevier Health Sciences, 2022.

\bibitem{glynn2022hutchison}
M.~Glynn and W.~M. Drake, \emph{Hutchison's Clinical Methods E-Book: Hutchison's Clinical Methods E-Book}.\hskip 1em plus 0.5em minus 0.4em\relax Elsevier Health Sciences, 2022.

\bibitem{joy2020multiclass}
M.~M.~H. Joy, M.~Hasan, A.~S.~M. Miah, A.~Ahmed, S.~A. Tohfa, M.~F.~I. Bhuaiyan, A.~Zannat, and M.~M. Rashid, ``Multiclass mi-task classification using logistic regression and filter bank common spatial patterns,'' in \emph{International Conference on Computing Science, Communication and Security}.\hskip 1em plus 0.5em minus 0.4em\relax Springer, 2020, pp. 160--170.

\bibitem{miah2022eeg}
A.~S.~M. Miah, M.~Hadiuzzaman, M.~S. Ali, and T.~M. Mahdee, ``Eeg-based hand gesture classification using machine learning approach,'' \emph{BAUST JOURNAL}, p.~19, 2022.

\bibitem{miah2019motor}
A.~S.~M. Miah, S.~R.~A. Ahmed, M.~R. Ahmed, O.~Bayat, A.~D. Duru, and M.~K.~I. Molla, ``Motor-imagery bci task classification using riemannian geometry and averaging with mean absolute deviation,'' in \emph{2019 Scientific Meeting on Electrical-Electronics \& Biomedical Engineering and Computer Science (EBBT)}.\hskip 1em plus 0.5em minus 0.4em\relax Ieee, 2019, pp. 1--7.

\bibitem{Breiman2001}
L.~Breiman, ``Random forests,'' \emph{Machine Learning}, vol.~45, no.~1, pp. 5--32, 2001.

\bibitem{dixi}
S.~Dixit, I.~Yekkala, and M.~A. Jabbar, ``Prediction of heart disease using particle swarm optimization,'' \emph{IEEE Int. Conf. Smart Tech.}, pp. 691--698, 2017.

\end{thebibliography}

\end{document}